\documentclass{article}

\usepackage{arxiv}

\usepackage[utf8]{inputenc} 
\usepackage[T1]{fontenc}    
\usepackage{hyperref}       
\usepackage{url}            
\usepackage{booktabs}       
\usepackage{amsfonts}       
\usepackage{nicefrac}       
\usepackage{microtype}      
\usepackage{lipsum}
\usepackage{graphicx}
\graphicspath{ {./images/} }
\usepackage[authoryear]{natbib}
\usepackage{tikz}
\usetikzlibrary{arrows.meta, positioning, fit, shapes.geometric}

\usepackage{tcolorbox}

\newtcolorbox{definition}[2][]{%
  colback=blue!5!white,
  colframe=blue!75!black,
  fonttitle=\bfseries,
  title=#2,
  #1}

\title{Systems Explaining Systems: A Framework for Intelligence and Consciousness}

\author{
	Sean Niklas Semmler \\
	Independent Researcher\\
	Master's Student at LMU Munich \\
	\texttt{sean.semmler@campus.lmu.de}
}

\begin{document}
\maketitle
\begin{abstract}
Understanding intelligence and consciousness requires moving beyond cataloging cognitive abilities to identifying the fundamental operations that produce them. This paper argues that intelligence emerges from a single purpose:forming, refining, and integrating causal connections between signals, actions, internal states, and learned structures. Through context enrichment—interpreting incoming information using previously learned relational context—intelligent systems achieve efficient processing under metabolic constraints. Rather than rebuilding environmental models from raw input for context, the brain reactivates stable causal structures learned from experience, enabling sophisticated interpretation from minimal sensory data.
Building on this foundation, we introduce the systems-explaining-systems principle: consciousness emerges when recursive neural architecture enables higher-order systems to explicitly learn and interpret the relational patterns of lower-order systems across time. Unlike simple stimulus–response networks that merely implement learned relations implicitly, recursive hierarchical systems can represent the underlying connections that give those relations meaning. Fed back through context enrichment and integrated within a dynamically stabilized meta-state, these interpretations transform the brain’s internal model from representing only the external world to representing its own cognitive processes.
The framework also reframes predictive processing theories. We embrace similar architectural principles but propose that prediction emerges as a consequence of a conceptually simpler underlying operation focused on discovering stable contextual structure rather than forecasting future inputs.
For artificial intelligence, this suggests progress toward human-like cognition may require multi-system architectures where higher modules interpret and regulate lower ones, rather than scaling single feedforward systems. Consciousness is not a separate faculty but the natural outcome of sufficiently deep recursive relational organization—intelligence turned inward upon itself.

This work is presented as a preprint, and the author welcomes constructive feedback and discussion.
  
\end{abstract}


\section{Introduction}
Traditional definitions of intelligence often enumerate the abilities humans display: reasoning, planning, abstraction, problem solving, learning, and adaptation (\cite{gottfredson1997mainstream, sternberg1986intelligence, halpern2014critical, deary2006genetics, gignac2024defining, zeigler-hill_conceptualizing_nodate}). Although descriptively useful, such lists do not identify the underlying mechanism that unifies these capacities. Psychometric definitions likewise frame intelligence as the “capacity to achieve goals” or to perform cognitive tasks successfully (\cite{gignac2024defining}). Yet these abilities represent outcomes of intelligence, not its working mechanisms.

This work begins from a different premise.
To understand intelligence, we must shift our perspective from viewing it as a collection of domain-specific skills to seeing it as a general process rooted in a single fundamental operation: forming connections. A system becomes intelligent when it can create, refine, and integrate connections between signals, internal states, actions, and previously learned structures. These connections form the relational architecture through which the system interprets the world, guides behaviour, and ultimately constructs meaning.

In earlier work (\cite{semmler2025context}), I introduced preliminary versions of the concepts of meta-state and context enrichment. That earlier formulation was exploratory and primarily focused on potential neurobiological mechanisms, emphasizing technical implementation details and energy-efficient neural dynamics. While this approach established biological plausibility, it lacked a unifying conceptual framework that clearly articulated what role these mechanisms play in intelligence and how they relate to higher-level phenomena such as cognition and consciousness.
The present paper represents a substantive shift in perspective rather than a purely incremental refinement. Here, the focus moves from mechanistic speculation to a conceptual account. The meta-state is defined not merely as a pattern of neural activity but as the system’s representation of environmental surroundings, and context enrichment is identified as the fundamental interpretive operation through which intelligent systems assign meaning to sensory input. Rather than emphasizing how these processes might be implemented at the neuronal level, this work clarifies their functional role in enabling flexible, context-dependent interpretation of their surrounding environment.
A signal’s meaning depends not primarily on its raw sensory properties but on the contextual structures through which it is processed. This shift in emphasis explains why identical inputs can yield radically different interpretations depending on the system’s current state and learned associations, and it situates context enrichment and the meta-state as central organizing principles rather than implementation-specific mechanisms.
 
Our framework builds substantially on recent hierarchical theories of cortical function, particularly Hawkins’ Thousand Brains Theory and broader predictive-processing frameworks (\cite{hawkins2021thousand, friston2010free, clark2013whatever, rao1999predictive, knill2004bayesian}). Mechanistically, our view shares many of the same architectural elements—contextual activation, pattern completion, and recurrent integration across cortical and thalamic loops. We embrace these architectural insights but offer a reframing of their functional interpretation. Predictive-processing theories typically describe cognition in terms of minimizing prediction error by forecasting future inputs. We do not reject this characterization. Instead, we propose that prediction emerges as a consequence of a conceptually simpler underlying operation: the discovery and reactivation of causal connections. What these frameworks call prediction, we interpret as context enrichment—the reuse of learned internal representations of the environment to determine the meaning of new sensory signals. The system provides the information the input alone can not provide. By shifting the emphasis from forecasting to relational interpretation, the Stable State Framework provides an alternative conceptual lens for understanding the roles of context, recursion, and hierarchical organisation in perception and cognition.

Building on these foundations, this paper introduces a second major contribution: the proposal that intelligence—and ultimately consciousness—emerges through systems explaining systems. Evolution progressively added layers of neural organisation in which higher systems do not simply connect incoming signals implicitly, but explicitly learn the implicit information expressed by lower systems. This idea resonates with higher-order theories of consciousness (\cite{rosenthal2009higher}), which hold that a mental state becomes conscious when it is represented by another system, and with global workspace or global neuronal theories (\cite{baars1997theatre, dehaene1998neuronal, mashour2020conscious}), which emphasise the widespread influence of integrated representations across the brain. In our framework, these insights converge: Recursion enables higher-level systems to develop internal activations over time, forming increasingly abstract and stable explanatory frameworks of both the external and internal environments. Crucially, these high-level frameworks are fed back into lower systems through context enrichment. Higher systems provide condensed, generalised representations of “what the organism is doing” and “what situation it is in,” allowing lower systems to process incoming signals more efficiently by interpreting them within this abstracted context. A low-level visual cue such as a shadow is therefore processed differently depending on whether higher systems represent the situation as “danger,” “play,” or “home.” This bidirectional flow—explanation upward, contextual guidance downward—gradually builds a sophisticated meta-state that models not only the external environment but also the internal organisation of the system itself.

This paper is conceptual in nature.
Its goal is not to propose new biological mechanisms, but rather to develop an organisational framework that encapsulates the functional principles that underpin intelligent behaviour in both biological and artificial systems.
The theory aims to provide an intuitive lens for understanding intelligence, clarify how meaning and abstraction emerge from simple relational operations, and offer a potential pathway for future artificial systems that move beyond current limitations.
By combining the ideas of context enrichment, meta-state dynamics, and systems explaining systems, this work proposes an updated framework for intelligence—one that is unified, relational, and capable of grounding both cognition and consciousness in the same underlying principles.

\subsection{Contributions}
\begin{enumerate}
	\item\textbf{ A refined theoretical foundation for intelligence}, framing it as the formation and integration of connections rather than prediction or task-specific abilities.
	\item \textbf{ A clearer and more rigorous formulation of meta-state and context enrichment}, replacing earlier definitions from \cite{semmler2025context} with precise conceptual foundations .
	\item \textbf{The introduction of the “systems explaining systems” principle}, proposing a recursive mechanism for the emergence of higher-order cognition and consciousness.
	\item \textbf{A unified relational architecture} that links intelligence and conscious processing within a single conceptual framework.
	\item \textbf{Implications for future AI design}, suggesting that progress may require multi-system architectures where higher modules interpret and regulate the representations of lower ones.
\end{enumerate}

The paper first develops the core components of the framework—connection formation, context enrichment, and the meta-state—and shows how they structure intelligent processing across different levels of complexity. It then introduces the systems-explaining-systems principle as the basis for higher-order cognition and consciousness, discusses implications for neuroscience and AI, and concludes with limitations and directions for future work.

\section{The Stable State Framework}

In this chapter, we introduce the conceptual building blocks of our framework. These include the notions
of context enrichment, the meta-state and the concept of frameworks, which together constitute the structural and dynamic core of intelligent
processing.

\subsection{Intelligence as the Capacity to Form Connections}

Every nervous system, even the simplest, performs one essential operation: it transforms signals into information.
A signal is a raw input—light, sound, pressure, internal drive.
By itself, a signal has no meaning. Meaning arises only when the system learns relations between signals, actions, and outcomes (\cite{harari2024nexus}).
 When a photoreceptor triggers a motor response, the organism has encoded a basic relational pattern: “when signal A occurs, action B should be taken.” The signal a has carried a certain kind of information for the organism. This is the earliest form of intelligence.
 In the Stable State Framework, we use the term connection to describe this general class of learned relations. A connection is not merely a synapse; it is a pattern that represents how two or more signals, states, or actions relate. A connection is therefore a unit of information. 
Throughout this paper, we use the terms meaning and information largely interchangeably. A signal carries meaning—or information—for an individual when it has formed a connection to that individual's internal states, actions, or other learned structures. Signals without such connections remain meaningless noise.
For example, a red light is just a signal; connecting it to a “danger” signal transforms it into a useful information.

As nervous systems evolved, their relational capacity expanded—from linking individual signals to linking internal states, behavioural outcomes, and eventually relations among relations. We therefore define intelligence as follows:
\emph{Intelligence is the capacity of a system to create, refine, and integrate connections between signals, actions, and internal states, and to use these connections recursively to guide behaviour.}
The formulation rests on three pillars:

\begin{enumerate}
	\item \textbf{Connection Formation.} Early organisms learned that certain sensory conditions cause beneficial or harmful outcomes. This is associative learning (\cite{wagner1972inhibition, pearce2001theories}), where the systems connects certain stimuli with actions, thus assigning an implicit information to that stimuli. Each learned connection turns a raw signal into meaningful information.
	\item \textbf{Meaning Assignment.} In biological organisms, signals gain meaning when they connect to internal states - affective, motivational, or homeostatic signals. Meaning is not an abstract label but reflects how a signal relates to the systems needs and well-being (\cite{damasio1999, thompson2010mind}). The same signal may hold different meaning depending on both the organism and also the internal state of an organism.
	\item \textbf{Recursive Integration.} More advanced forms of intelligence arise when systems form connections between existing connections. This recursive capability enables the construction of sophisticated hierarchies, abstractions and concepts (\cite{tenenbaum2011grow}). 
\end{enumerate}
 Crucially, connection formation is not only a mechanism for learning but also for selecting relevance. The environment is saturated with signals, most of which are irrelevant to a given organism. A signal that is not connected to internal states, actions, or outcomes carries no meaning for the system and is ignored without consequence. Only when a signal becomes part of a learned connection—linked to reward, danger, goals, or behavioural consequences—does it turn into information.
 
These operations transform raw sensory input into an organised connected world. Intelligence, in this view, is not prediction in the narrow sense, nor a collection of domain-specific skills. It is the ability to construct and use meaningful relations, and to do so recursively across time and levels of abstraction. All complex cognitive phenomena—generalisation, planning, inference, imagination, and consciousness—emerge from this fundamental operation.

\subsection{Context Enrichment as the Fundamental Processing Mechanism}

A fundamental challenge for any intelligent system is identifying meaningful signals within an overwhelming stream of sensory input while operating under strict metabolic and computational constraints (\cite{attwell2001energy, harris2012energy}). Neural activity is energetically expensive, consuming approximately 20\% of the body's energy despite representing only 2\% of body mass (\citep{attwell2001energy, harris2012energy}). Because sensory bandwidth and neural resources are finite, processing every signal anew is energetically prohibitive. Biological evolution solved this not by expanding sensory capacity, but by developing architectures that enrich incoming signals with context derived from past experience. Meaning is not found solely in raw data; it is constructed by interpreting data through a dense network of previously learned information.

The sophistication with which a system interprets a signal depends not only on the signal itself, but on the context in which it is processed.
Simple organisms respond only to immediate input: a light flash always triggers the same reaction because no additional contextual information is available to modulate the response. More intelligent organisms interpret signals against broader situations—location, timing, emotional relevance, past associations. Humans represent the extreme: the same sensory input can evoke radically different meanings depending on contextual situation. The difference lies in how much of the surrounding environment and internal state is taken into account when interpreting a signal.
However, processing the entire environment for each incoming signal would be prohibitive. Instead, intelligent systems rely on learned, compact representations of their environment that persist through recursive neural dynamics. The context used for processing does not need rebuilding from scratch; most remains stable, requiring fresh processing only for changing details. When a system has learned a stable relation between a signal and a contextual framework, interpretation becomes efficient: the relevant relational structure is already active.

To describe these learned, compact representations that supply context for processing of signals, we introduce the term framework.
A framework is a conceptual structure that serves as a cluster of connected information to model an aspect of the world or in the system’s own activity. In this sense, a framework functions similar to a model of an object (\cite{hawkins2021thousand, lecun2022path, albus2002outline})—but one that is actively used rather than passively stored.
We choose the term framework over model to emphasize this functional role. A framework is not merely a descriptive representation; it is the platform through which signals acquire meaning. When a signal arrives, it is interpreted relative to the currently active frameworks, which determine what aspects of the signal are relevant and how it should influence behaviour.
Frameworks and their relation to models, abstraction, and the meta-state are discussed in more detail in Section \ref{sec:model}

Crucially, this contextual structure does not only enrich meaning—it also filters information. Active frameworks provide generic explanations for most incoming signals, allowing them to be processed with minimal effort. Signals that are consistent with the current context are effectively “explained away” as expected background and do not require further elaboration. Only signals that conflict with, extend, or are especially relevant to the active framework demand additional processing. In this way, context enrichment serves as a powerful relevance filter: it compresses the vast majority of sensory input into stable, low-cost representations while selectively amplifying signals that matter for current goals, threats, or learning. This allows intelligent systems to focus resources where they are needed without explicitly evaluating every signal in isolation.

This capability relies fundamentally on recursion. Before the evolution of brain-like architectures, neural systems were largely limited to simple stimulus–response connections, processing each signal anew and learning only from events that occurred simultaneously.  In its simplest form, connecting new inputs means linking two signals to represent “these belong together and result in the same output.” Context was only what the system could receive at the same time. With the emergence of recursive structures, this changed fundamentally. Recurrence allows previous activations to persist. This persistence forces the system to link current input with past activity, enabling the formation of stable, generalizable and abstract representations of the systems environment.

We term this process context enrichment. When a new signal arrives, it is immediately integrated into the system’s existing relational network—its frameworks—using prior context as a scaffold.
If the connection between a current input and a previously learned structure already exists, the system can interpret it with minimal processing. If such a connection is missing, it must first be discovered — a process we will examine in detail in chapter \ref{sec:learning}.
Context enrichment functions through a bidirectional dynamic:

\begin{itemize} \item \textbf{Prior structure shapes new signals:} Incoming events are interpreted via currently active frameworks. A sound acquires specific meaning (e.g., danger vs. novelty) depending on the pre-existing state of the system. \item \textbf{Current input reshapes prior structure:} When an event conflicts with or extends existing connections, the relational network reorganizes. Experience does not overwrite old patterns but reconfigures them, producing a richer internal model. \end{itemize}

This mechanism closely resembles the computational steps described in predictive-processing frameworks, and in fact relies on many of the same neuronal principles—contextual activation, pattern completion, and recurrent stabilisation (\cite{hawkins2021thousand, friston2010free, clark2013whatever, knill2004bayesian, rao1999predictive}). However, we use the term context enrichment to emphasise a different conceptual interpretation. Rather than viewing the brain as primarily anticipating future inputs, we propose that the core function is to use previously learned causal connections to enrich and guide the processing of current input. When an incoming signal arrives, the recursive architecture has already activated neurons representing the context of that signal, thereby providing additional information that makes processing more efficient. If the connection between the current context and the signal has already be learned, then the processing will activate similar stable representations again. What appears as "prediction" is thus not a dedicated computation aimed at forecasting the future, but the natural consequence of the system reactivating learned causal mappings that relate past, present, and potential future events. The underlying mechanism is the same, but the conceptual interpretation shifts: than characterizing the brain as attempting to forecast future inputs, we propose that it reuses learned causal structure to efficiently interpret current inputs. Prediction-like behavior emerges as a consequence—when context activates learned connections, the system naturally facilitates processing of causally related events—but prediction is not the primary computational goal. In that sense, context enrichment is broadly compatible with predictive-processing mechanisms, but reinterprets their functional role: not as a system optimising future predictions, but as one optimising present interpretation under resource constraints. This reframing has implications for how we understand learning priorities: rather than minimising prediction error per se, the system seeks to discover stable causal relations that efficiently connect elements of experience.

Consequently, intelligence is tightly linked to the richness of context a system can draw upon. Systems lacking recursive architecture must rely on immediate input, requiring high data density to find patterns. They can only detect simple patterns and require more raw information to differentiate between situations. Recursive systems, by contrast, reuse prior activations as contextual scaffolding, enabling far more sophisticated interpretation and meaning-making from minimal sensory cues.
The more prior structure is available — accumulated experience, abstractions, higher-order frameworks — the finer the distinctions the system can make between signals that may appear nearly identical at the sensory level.
 This explains why experts perceive vastly more meaning than novices when viewing the same object: their internal frameworks provide a richer structural scaffold for interpretation. 
 As we will see in later sections, most of this context is generated by higher-order systems that learn the structure of lower systems. These higher systems provide the interpretive frameworks that shape how signals are processed, enabling efficient connection formation and ultimately grounding meaning, abstraction, and consciousness.

\subsection{Causal Connections as the Fundamental Learning Unit}
If context enrichment describes how intelligent systems process information efficiently, we must ask: what do they learn? We propose that the fundamental unit of learning is not a probabilistic but a causal connection between structured patterns.

In biological organisms, learning is often influenced by internal states such as reward, punishment, comfort, or discomfort. When a connection is consistently associated with a positive outcome, neuromodulatory signals strengthen it; when associated with negative outcomes, it weakens (\cite{schultz2015neuronal}). This is the foundation of reinforcement learning, perhaps the most ancient and robust form of adaptive behaviour (\cite{niv2009reinforcement, dayan2008decision, barto1997reinforcement}).
 At the neural level, this means: when neurons A and B activate in temporal proximity, neuron C learns to activate in response, encoding "A and B together lead to C." Behaviourally: "When pattern A occurs, we should perform action X to receive outcome B." \\
The temporal asymmetry of neural activation naturally supports the learning of causal order. While often summarized as “neurons that fire together wire together,” Hebbian learning is more accurately understood in its temporally sensitive form: neurons that fire in reliable sequence develop directional connections (\cite{hebb1949organization}). As a result, the system does not merely learn that events co-occur, but that one event tends to precede and give rise to another. This directional structure is what makes learned connections behaviourally useful. Functionally, neurons and their synaptic connections participate in representing causal relations between events, albeit implicitly and in a distributed manner. Recurrence extends this capacity by allowing earlier activations to persist long enough to be associated with later outcomes, enabling the system to learn causal structure explicitly across time rather than only instantaneous correlations.
Hierarchical organisation then allows higher cortical regions to integrate sequences of causal connections learned at lower levels. A higher-order framework may encode an entire behavioural sequence—reach → grasp → lift—while the individual causal steps are represented and learned by lower regions. Because higher frameworks are more stable, they can guide the activation of their constituent causal links in the correct order, orchestrating coherent behaviour without re-learning each step during execution.

Seen from this perspective, prediction is not the fundamental learning operation. The brain does not begin by attempting to forecast future sensory input. Rather, prediction emerges naturally from the causal connections that have already been learned. When the A-part of an A→B relation becomes active, the system helps to activate the B-side—not because it has computed a probabilistic estimate, but because the context makes B easier to activate. What appears as prediction is thus context enrichment applied to a learned causal structure: the brain activates the relational pattern that describes the current environment, helping the system to process current input, which was already learned in that context and can be recognized easily.
In essence, the brain learns causal connections—what appears to observers as 'prediction'—and continuously reactivates them during processing. However, it does not learn through prediction; rather, it learns through discovering and strengthening causal connections, with prediction-like behaviour emerging as a consequence. 
When signals and active context have no learned connection, the system must modify either signal interpretation or context representation—typically by gathering additional information through attention.

Finally, this account lays the foundation for understanding consciousness. While simple organisms can learn implicit causal relations—“A leads to B”—recursive and hierarchical architectures enable organisms to represent these causal structures explicitly, to integrate them across contexts, and eventually to understand their own actions in relational terms. As later sections (\ref{sec:consciousness}) show, the shift from implicit causal sensitivity to explicit causal interpretation is a critical step in the emergence of conscious cognition.

In summary, learning is best understood as the search for efficient causal structure, driven by metabolic constraints (\cite{attwell2001energy, harris2012energy}). The brain seeks not perfect prediction, but efficient interpretation: the ability to process incoming signals through a rich internal network of learned causal relations. Prediction, when it occurs, is a natural consequence of this relational architecture—not its foundation.

\subsection{The Meta-State: The Coherent Structure of an Internal World}
Having established that intelligence operates through context enrichment of causally structured connections, we can now define the system's active context more precisely. At the heart of the Stable State Theory is the concept of the meta-state—the integrated representation of all active and latent relational structures available to the brain at a given moment. The meta-state is not a static representation; it is a dynamic, self-organizing pattern that adjusts continuously as new information is incorporated.
\begin{definition}{Meta-State}
The concept of the meta-state is defined as the distributed pattern of all neural activity across the entire brain. This pattern represents contextual information, including physical location, emotional state, goals, and prior knowledge at a certain point in time. It serves to enhance processing efficiency for contextually relevant inputs while inhibiting processing of irrelevant information. 
\end{definition}

It is critical to understand that the meta-state is not confined to any single circuit or region of the neocortex but emerges from the integrated activity of the entire brain, including subcortical structures, brainstem nuclei, and all cortical layers.
The meta-state includes current perceptual input, relevant memories and conceptual structures, emotional and motivational states, and the system’s model of itself in relation to the environment. 
The meta-state is the persistent environmental representation that makes context enrichment possible—the active scaffold through which incoming signals are interpreted.

The meta-state must balance two competing demands.
First, stability: It must be able to support the processing of diverse incoming signals without collapsing or reorganizing entirely for each new input. This requires that internal representations can be reactivated repeatedly, even when the external input varies slightly. We propose that this stability is supported by circular activation dynamics within and between brain regions (\cite{salners2023recurrent, rajan2016recurrent}). The cortico-thalamo-cortical loop plays a special role. Higher cortical regions send feedback signals to the thalamus, which then shapes how incoming sensory input is filtered and relayed back to the cortex (\cite{sherman2016thalamus, hawkins2025hierarchyheterarchytheorylongrange}). Through this loop, the thalamus can bias sensory processing toward patterns that match the currently active cortical state. As a result, even when the sensory input changes slightly, the thalamus helps reproduce similar activation patterns in the cortex, preserving the meta-state coherence. In this way, the system uses prior internal representations as feedback to stabilize ongoing activity. Inputs that “belong together” in meaning are guided toward similar cortical activation patterns, enabling the meta-state to remain robust while still flexible to new information. \\
 Second, flexibility: Despite its need for stability, the meta-state must remain responsive to new information. New inputs must be allowed to change parts of the meta-state slightly, so that an individual can adapt to new situations. The brain's hierarchical organization achieves this balance: lower-level regions change activity rapidly to track moment-to-moment sensory variations, while higher regions update more slowly, maintaining stable, abstract representations persisting across changing inputs. As information ascends the hierarchy, representations become increasingly stable and integrative, yet still draw on fine-grained dynamics of lower regions. This combination of fast, flexible signals and slow, stable frameworks enables the meta-state to remain coherent while adapting continuously.

The meta-state is the platform upon which intelligent behaviour operates. All perception, memory, decision-making, and learning arise within its structure. It addresses intelligence's fundamental challenge: processing information with limited metabolic resources. Representing every environmental object with dedicated neural activity would be prohibitively costly. Instead, the meta-state provides the efficient contextual scaffold that makes sophisticated interpretation possible from minimal sensory input.

\subsection{Models of the world} \label{sec:model}
The meta-state can be understood as the system’s currently active portion of a much larger internal model of the world. Similar ideas appear in several influential theories (e.g., \cite{hawkins2021thousand, lecun2022path, albus2002outline}), all of which assume that intelligent behaviour requires some form of structured internal representation.
Whether one views this internal model as many smaller models or as one integrated model is secondary. For our purpose, the meta-state reflects the active subset of a single, large relational model built over the system’s entire learning history.
 We describe the nature of this internal model as networks of learned connections linking sensory signals, actions, internal states, and previously constructed abstractions.

We use the term framework (also abstraction) to describe the interconnected relational clusters that make up this model. Frameworks are similar to what Hawkins (\cite{hawkins2021thousand}) calls “models”: coherent relational clusters of connections of different information or signals.
We adopt the term framework rather than model because it better captures their functional role in our theory—frameworks serve as platforms on which incoming information is interpreted and integrated into the system’s relational structure.

A framework is not a passive representation. It is an active processing substrate: a relational scaffold that determines how new information is understood and where it fits within the system’s existing knowledge. When a new signal arrives, it is connected to a relevant framework, which provides both the contextual meaning and the potential linkage points to other structures. Frameworks do not need to be completely accurate representations of the real world. They only need to be stable and useful enough to guide behaviour.

Frameworks range from concrete (a particular face, object, place) to compositional (“dog,” composed of sub-frameworks like “animal,” “four-legged,” “loyal”) to highly abstract (justice, growth, democracy). Most frameworks are themselves sets of connections between other frameworks.
This mirrors Hawkins’ observation that cortical columns learn compositional models—objects built from other objects and features (\cite{hawkins2021thousand}). Just as a bicycle is composed of wheels, frame, pedals, and seat, or a word is composed of syllables and letters, a framework such as “dog” is composed of many interconnected sub-frameworks that capture its relational essence. 
Crucially, in both Hawkins’ theory and ours, hierarchy is not about progressively larger features, but about compositional structure. Higher regions represent more abstract frameworks composed of those learned by lower regions. The hierarchy reflects the integration of components into wholes, not the simple accumulation of complexity. As Hawkins notes, the neocortex does not need to represent all compositional levels simultaneously—only the relevant subset at any given moment. The same principle applies here: frameworks provide the structure through which multiple layers of meaning can be accessed flexibly and efficiently. 
As frameworks become more abstract, they encode a greater number of relational connections, serving as hubs that can activate diverse associated structures depending on context. This expanded connectivity enables rapid interpretation, generalization, and efficient meaning-making across a wide range of sensory and internal conditions.

When discussing neural mechanisms, we also often use the term pattern. A pattern is the neural activation configuration that reflects the system having discovered a connection. When neurons detect that two signals consistently occur together—through temporal proximity, shared population activation, or contextual co-occurrence—they form a connection, which manifests as a stable pattern of activation.
Thus:
\begin{enumerate}
	\item Connection: learned relation between two signals or structures
	\item Pattern: neural activation that encodes that relation
	\item Framework: a coherent cluster of many such relations
\end{enumerate}
 
This architecture explains why humans can maintain rich conceptual worlds despite limited cognitive capacity. Abstractions reduce the need to store all details explicitly; instead, they organize knowledge so that details can be reconstructed or retrieved when necessary. This yields dramatic efficiency. For instance, people can recall the plot of a complex narrative because the story provides a pattern of meaningful relations—each event is connected to others, constraining interpretation. In contrast, random digits lack relational structure; they require the system to maintain each item independently, without the benefit of shared connections.

We propose that a framework is sustained in the meta-state through recursive activation patterns, as reported by (\cite{salners2023recurrent, rajan2016recurrent}). A framework corresponds to a particular pattern of neural activation of other systems—for example, the sequence of actions required to perform a task. As this sequence unfolds, each step activates the neurons encoding that part of the sequence. These activations, in turn, reactivate the framework’s core representation, which then provides the contextual structure needed to maintain the meaning and coherence of the task. In this way, ongoing neural activity and the framework’s relational structure reinforce each other, keeping the framework active as long as it remains relevant.

\section{Learning and memory} \label{sec:learning}

Learning and memory have traditionally been viewed as distinct cognitive processes. We propose that they are not separate faculties but two expressions of the same underlying process: the formation, stabilization, and organization of connections and thus the creation of frameworks.  An intelligent system accumulates experience by discovering connections—between signals, actions, internal states, and previously learned structures—and transforming these connections into enduring frameworks that support flexible behaviour. We defined a connection as a learned relation linking signals, actions, internal states, or learned frameworks, thus creating an information. By accumulating such relations over time, the system constructs increasingly rich internal frameworks that allow it to interpret new experiences, to generalize across contexts, and to guide adaptive behaviour.

In this chapter, we outline how connections are formed, reinforced, abstracted, and organized into hierarchical relational layers. We also examine how working memory and long-term memory interact, how attention functions as a mechanism for discovering missing connections, and how the evolutionary layering of cognitive systems contributes to higher-order intelligence. Our discussion is intentionally conceptual. We do not aim to specify detailed neural mechanisms or learning algorithms, but to identify the functional principles that govern how learning is organized across levels of abstraction. Many implementation-level questions are beyond the scope of this paper and are left for future work.

\subsection{Learning as the Discovery of Connections}

At its core, learning is the process of discovering reliable relations across experience. When two signals consistently co-occur, when an action repeatedly produces a certain outcome, or when internal states align with specific behaviours, a neural systems forms a connection linking these elements. This is the fundamental principle underlying associative learning (\cite{wagner1972inhibition, pearce2001theories}) and reflects the basic principle articulated by \cite{hebb1949organization}: elements that activate together become structurally linked.

In biological organisms, the discovery of relationships is fundamentally based on the recognition of patterns in neural activation. A connection is formed when the system detects consistent patterns in the activation of neural populations. The ability to recognise patterns is the most fundamental cognitive skill of the brain. Neurons are built to recognise patterns in the activation of other neurons. In the simplest form, it detects whether the previous neuron has been activated. In a more complex setup, it recognises a pattern of activation in a set of neurons, thus 'connecting' them (\cite{rosenblatt1958perceptron}). The more often a subset of neurons in the set fire, the stronger the connection becomes and the less additional input from other neurons is necessary .  

In purely feedforward systems, this learning remains limited to immediate associations: a particular input triggers a particular response because, over evolutionary or developmental time, this response has proven beneficial. The causal relation between the input, the action, and the eventual outcome remains implicit—encoded only in the strength of the connection itself.
Recursion fundamentally changes this situation. By allowing neural activity to persist across time, recursive circuits enable the system to bind temporally separated events into a single representational structure. An input can remain active while an action is executed and while its consequences unfold. As a result, the system can learn not merely that a response follows a stimulus, but that performing a particular action in a particular context leads to a specific outcome. Causal relations—such as “in situation X, action A results in reward R”—become explicitly learnable structures rather than implicit implementation.

This capacity to represent causal structure across time marks the transition from simple associative learning to genuinely explanatory learning. It allows the system to model not only correlations in experience, but the consequences of its own actions, laying the foundation for goal-directed behaviour, planning, and later, conscious interpretation.

Through these pattern recognition processes, the system does not require explicit rules or external labels to identify relationships. The patterns in neural activity reflect the patterns of the world, and synaptic plasticity transforms these patterns into stable relational structures.

Over time, these primitive relations accumulate, enabling the system to interpret new signals not in isolation, but in terms of a developing relational network. When multiple frameworks are active simultaneously, new relations can form between the frameworks themselves. These higher-order connections generate abstract categories, concepts and generalised patterns that transcend specific sensory inputs.

Learning, in this framework, is the expansion and refinement of the system’s relational structure—the ongoing process by which previously unconnected elements become meaningfully linked. All advanced cognitive abilities emerge from this foundation.

\subsection{Attention as the Search for Missing Connections}

Learning occurs most strongly when the system encounters uncertainty—when incoming signals or internal states have no or only weak learned connection within existing frameworks. Uncertainty indicates that the system lacks the necessary connections to interpret or respond effectively. The mechanism that resolves this uncertainty is part of what we call attention.

When uncertainty arises, it is reflected in the brain by unusual high neuronal activity in the according regions (\cite{posner1989attention, corbetta2002control, gottlieb2013information}).
This is caused because there are no clear learned connections between the inputs and the activated frameworks. Because the additional information of the frameworks is now missing, neurons can not indefinitely tell which kind of pattern they are looking at, and thus multiple different patterns activate. Those over-activation serves two purposes:
Firstly, it draws our attention to the source of the uncertainty. As many parts of the brain influence movement, each of these over-activated areas can prompt an organism to move towards the source of the activation. Through this focus, the brain tries to gather more information to compensate for the lack of internal information. The more information acquired, the easier it is to recognise patterns and find connections between them.
  Second it activates additional relational frameworks. Activating multiple patterns is identical to activate multiple frameworks that could be used for processing. This is the system's search for a plausible context. These activated, detailed frameworks represent a larger set of pre-existing neurons that the new signal representation can potentially connect to. In essence, attention is a directed search: it not only deepens the target signal (creating more connection points) but also broadens the search space of potential targets (creating more connection points). This strategic deployment of cognitive resources is what ultimately resolves uncertainty by discovering missing relations and expanding the system's internal structure.

\subsection{Working Memory as the Active Meta-State}
The working memory (often inconsistently called Short-Term Memory) and its interaction with the Long-Term Memory (LTM) has been substantial part of brain research (\cite{baddeley2020working, baddeley1983working, christophel2017distributed, barbosa2020interplay}). We do not presume to provide a complete explanation of the functionality of these systems and the underlying neural details. That is beyond the scope of both this work and our expertise in this field. Instead, we want to propose a rough conceptual idea of how working memory and LTM can be viewed in the context of stable state theory. 
The working memory corresponds to the currently active meta-state, defined through current activated frameworks. Its capacity is severely limited, reflecting biological constraints on maintaining a large number of simultaneous, highly-active neuronal connections (\cite{baddeley2012working, cowan2005capacity}).

Yet this limitation is mitigated by the structure of the frameworks themselves. A single abstract framework may encapsulate many connections. Experts, who have developed sufficient abstract frameworks, can hold complex sequences of tasks in their working memory. Novices, lacking such unifying frameworks, must treat each item as a separate framework, quickly exhausting their capacity.

The significance of working memory lies not in its capacity but in its function as the brain’s workspace for relational discovery. When frameworks are simultaneously active within the meta-state, their overlap creates opportunities for detecting novel associations between them. Crucially, the brain’s recursive architecture allows these activations to persist over time, even when the original sensory input has faded. This persistence keeps representations active long enough for the system to process them with new inputs or with other ongoing activations. As a result, the system can learn relations between events that unfold across extended time spans—relations that would be impossible to detect in a purely feedforward or moment-to-moment processing scheme. Through repeated co-activation, these transient relations become strengthened, establishing more efficient neuronal representations and stabilizing the newly discovered connections for long-term retrieval.

\subsection{Long-Term Memory as Stabilized Frameworks}

Long-term memory (LTM) is the stable substrate of the relational network. In the literature, there is ongoing debate about whether short-term memory (STM) and LTM arise from the same underlying mechanism (\cite{cowan1988evolving, nairne2002remembering, neath2009short, norris2017short}). For our purpose, we treat them as the same system.
 It is in essence the result of developing long-lasting connections between signals and information. 
Working memory provides the active workspace in which new connections are created. It maintains currently relevant frameworks—both detailed and condensed, abstract ones—so that they can guide interpretation and be reactivated when needed. By keeping these representations active, working memory exposes them to ongoing neural activity, enabling the system to detect recurring patterns and to strengthen the synaptic links that encode them. As  \cite{hebb1949organization} described, repeated co-activation leads to more robust and long-lasting connections.
Whether a connection may be classified as “short-term” or “long-term” depends on two factors: Its stability and longevity, and how much contextual support it requires for activation.
New connections typically depend on the same contextual conditions under which they were formed; they must be “reminded” by the original constellation of signals or frameworks. As these connections strengthen through repeated activation, they become increasingly independent of specific contexts and can be triggered more easily. In this way, STM gradually transforms into LTM: the same mechanism operating at different degrees of stability and contextual dependence.

Learning emerges from the continuous interaction between working memory and long-term memory.
Working memory activates relevant frameworks, integrates new signals, and searches for connections.
Long-term memory constrains interpretation, offering existing relational patterns that guide understanding.
Learning occurs when repeated co-activation leads to the stabilization of new connections from WM into LTM.

\section{Consciousness: Systems Explaining Systems} \label{sec:consciousness}

Consciousness is often treated as a mysterious or irreducible property of biological systems, something fundamentally different from ordinary cognition. However, we propose that consciousness arises naturally from the architectural principles of intelligence. It is not an additional faculty layered on top of cognition; nor is it an immaterial essence. Instead, consciousness emerges when a system becomes capable of learning, representing, and interpreting the implicit causal connections of other systems—including its own.

This view aligns with several influential theories that define consciousness as the ability of a system to access, model, or understand its own internal states. Higher-order thought theories (\cite{rosenthal2009higher}), global workspace models (\cite{baars1997theatre}), and the attention schema theory (\cite{graziano2013consciousness}) all emphasize that consciousness involves awareness of one’s own processes, especially the ability to represent the causes and consequences of one’s own actions. In its simplest form, consciousness therefore can be understood as an understanding of one's own actions and internal states. In other words, it is the explicit awareness that one's behaviour has consequences.

In this sense, consciousness is the pinnacle of intelligence: the point at which a system no longer merely reacts to patterns but understands the patterns underlying its own activity. This capacity does not appear abruptly but unfolds gradually as organisms evolve additional systems. Consciousness, therefore, is best understood as a spectrum of relational depth, not a binary state.

The purpose of this chapter is to formalize this view and to show how consciousness arises from hierarchical relational organization, how it interacts with the meta-state, and how it enables meaning, intention, and flexible behaviour.

\subsection{From Reaction to Interpretation: The Basis of Consciousness}

At the lowest level of biological organization, nervous systems operate through direct connections: sensory patterns trigger fixed actions. The earliest nervous systems consisted of simple networks linking sensory inputs directly to motor outputs. A photoreceptor might activate a motor neuron, causing an organism to move toward or away from light (\cite{arendt2008evolution, erwin2015early, arendt2016origin}). These systems implemented the most primitive form of intelligence: the ability to connect a signal to an action. They behave, but they do not understand.  Their reactions are driven by learned or innate connections that encode relations like “pattern A → action B,” without any capacity to represent why this pattern existed.

Importantly, these early A → B connections implicitly carried evolutionary value: performing B in situation A had, historically, led to beneficial outcomes. But the organism could not represent that outcome as the result of its own action. The reward state R that followed action B was simply a new situation—not internally connected to A or B in any representational sense. The information carried by that signal-action connection remained implicit, encoded only in the connection strength itself, shaped through evolutionary pressure over generations. When such a system encounters a dark shadow while foraging, it flees—not because it "knows" shadows indicate danger, but because organisms that fled from shadows survived more often than those that didn't.

A profound transition occurred with the emergence of recurrent neural connections. Recurrence allows the system to keep previous activations partially active while new states unfold. This architectural change enables the system to integrate temporally separated events, discovering the causal connections between them. For the first time, the organism can represent situation A, action B, and resulting state R together, in a single temporal window.
Recurrent Processing Theory similarly identifies recurrence as critical for consciousness, showing that visual awareness requires recurrent feedback loops beyond initial feedforward processing (\cite{lamme2006towards, lamme2010neuroscience}). We extend this insight: recurrence enables not just awareness of current patterns, but the binding of temporally separated events into explicit causal structures. When a shadow appears, the system's state includes both the new sensory pattern and persistent contextual activations (forest, alert state, recent sounds). This allows learning context-dependent mappings: the same input produces different outputs depending on active context. The system discovers that context determines outcome.

A simple example illustrates this shift. Encountering a shadow in a forest, a lower system detects the sensory pattern and, through learned sequential associations, activates a flight response. A recurrent system, however, can encode not only the sequence “shadow $\rightarrow$ flight $\rightarrow$ safety” but also how this sequence depends on broader contextual patterns (dense forest, isolation, heightened arousal).

The next evolutionary leap occurred when higher-order neural systems emerged that could observe and learn the causal relations expressed through lower systems. These higher systems receive internal activation patterns rather than external stimuli and can represent relations between different lower systems' states, actions, and outcomes.

This architectural transition echoes the central insight of Higher-Order Thought (HOT) theories (\cite{rosenthal2009higher}): a mental state becomes conscious when the system has a representation of that state, rather than merely implementing it. In HOT theories, a first-order visual perception becomes conscious when accompanied by a higher-order thought about that perception. In our framework, a lower system's implicit causal mapping becomes explicitly represented when a higher system learns that mapping by observing lower system's activation patterns. We ground this in architectural observation: higher systems literally receive activation patterns from lower systems as input, learning their structure and dependencies through the same connection-forming mechanisms that underlie all intelligence.

The explanatory capacity of a higher system derives from its connectivity—specifically, from receiving inputs from multiple specialized lower systems simultaneously. A recurrent visual system, no matter how sophisticated, can only learn the temporal dynamics of visual patterns. But a higher cortical area receiving convergent input from visual cortex, motor cortex, limbic structures, and subcortical reward systems can learn something fundamentally different: how these different systems' activations relate to each other and to outcomes.
When our foraging animal encounters a shadow, the visual system detects the pattern, the amygdala evaluates threat, motor cortex prepares flight, and the hypothalamus modulates arousal. A higher system observing all these activations simultaneously can learn their causal structure: "When visual-pattern-X co-occurs with amygdala-threat-signal and motor-preparation, safety typically follows." This cross-system integration—learning how different specialized systems' outputs relate—is what enables explanation.

Through reinforcement mechanisms, effective mappings are strengthened. However, the lower system often cannot represent why the mapping works—it may not have direct access to reward signals, or may process them only as modulatory influences on connection strength. A higher system, receiving both the lower system's activation patterns and reward signals from subcortical structures, can represent the causal relation explicitly: "In this context, this input leads the lower system to produce this action, which produces this reward." This is the first explicit model of one's own behavior. The system does not simply flee from shadows—it represents why fleeing occurs: because shadows in certain contexts predict threat, and fleeing in those contexts produces safety. Implicit evolutionary information becomes explicit relational structure.

The architecture becomes truly intelligent and conscious when these higher-level interpretations are fed back down through context enrichment. This mirrors the broadcasting principle of Global Workspace Theory (\cite{baars1997theatre, dehaene1998neuronal, mashour2020conscious}): a representation becomes conscious when it is made globally available to guide ongoing processing across multiple systems. In Global Workspace Theory, a representation enters consciousness when it gains access to a "global workspace" and can influence diverse cognitive processes. In our framework, higher systems' learned causal structure is broadcast downward through context enrichment, shaping how lower systems process new inputs. The functional consequence is similar—learned structure becomes globally influential—but the mechanism differs: rather than a dedicated workspace, hierarchically organized systems naturally broadcast their learned interpretations through descending connections.

Higher systems supply abstract frameworks—"this environment is dangerous," "we currently searching for food on unknown terrain "—that enrich lower systems' processing. The same shadow now produces different behaviour depending on which higher-order frameworks are active. Context enrichment creates a recursive loop: lower systems implement causal mappings, higher systems learn and interpret those mappings by observing multiple systems simultaneously, and these interpretations are broadcast back down to shape ongoing processing.

This recursive architecture transforms the meta-state from a model of the external environment into a model of the system itself. In simpler organisms, the meta-state primarily represents external context: location, threats, resources, current needs. But as hierarchical systems accumulate, the meta-state increasingly represents internal processing. A higher system that has learned "when I'm performing complex task X, I progress through steps A→B→C" can maintain this task-framework as active context. The lower systems executing individual steps no longer need to represent "which step am I on?"—this information is provided as context from above. The higher system models the task structure; the lower systems implement the operations within that structure.
This internal modeling is what enables increasingly sophisticated behavior. A system that can efficiently represent its own multi-step processes can execute them more reliably, interrupt and resume them, chain them into longer sequences, and adapt them to varying conditions. The meta-state becomes not just "where I am and what's around me" but "what I'm doing, why I'm doing it and what stage I'm at." Intelligence grows as the meta-state becomes a more comprehensive model of both external and internal structure. And consciousness, correspondingly, deepens: the system doesn't just respond to the world, but represents its own processing within that world.

As hierarchical layers accumulate—each explaining the patterns of those beneath by integrating across multiple systems, each providing context back downward—the system gains progressively deeper interpretive structure. A still higher system may learn patterns such as: "I tend to flee in forests because my threat-system becomes active, and this activation is usually accurate in dense environments but often false in open areas." This system represents not just the behavior-outcome mapping, but the reliability of its own intermediate processing—its threat-evaluation system's accuracy in different contexts. This connects to the broadcasting principle of Global Workspace Theory in a precise way. When high-level systems—particularly those in prefrontal cortex that observe and integrate across many specialized systems—form abstract frameworks, these frameworks become part of the active meta-state. Because these frameworks represent the causal structure of multiple lower systems simultaneously, they can provide context that influences virtually all ongoing processing. A prefrontal representation like "I am in a high-stakes situation requiring careful deliberation" doesn't just describe the current state—it actively shapes visual attention, emotional regulation, motor inhibition, memory retrieval, and reasoning processes through context enrichment.
This is functionally equivalent to GWT's broadcasting, but mechanistically grounded in our framework: high-level frameworks that represent many systems can influence many systems, because context enrichment naturally propagates their structure downward through the hierarchy. What makes a representation conscious is not access to a separate "global workspace," but rather: (1) it is maintained in the meta-state at a high level of abstraction, (2) it integrates information across multiple specialized systems, and (3) through this integration and abstraction, it can provide context that shapes processing across those systems. Consciousness is the active presence, in the meta-state, of frameworks that model the system's own internal structure comprehensively enough to guide its own processing globally.
This is not just behaviour; it is meta-behavioural understanding—consciousness emerging naturally from the recursive application of the same connection-forming mechanisms that underlie all intelligence. The progression is continuous: from meta-states that model only external context, to meta-states that model simple internal processes, to meta-states that model the relationships between internal processes, to meta-states that model the modeling itself. Each level of sophistication represents both increased intelligence (more adaptive, context-sensitive behaviour) and increased consciousness (deeper understanding of one's own processing). As we will see in the next section, this architecture scales: systems can explain systems that explain systems, producing the increasingly abstract and reflective awareness characteristic of more complex organisms.

\subsection{Higher-Order Systems: From Environmental Models to Self-Models}

As we have seen, intelligent systems interpret new inputs through internal structure—the meta-state—which encodes relevant aspects of the current situation. Even in simple recurrent systems, this internal structure functions as an implicit model of the environment, representing location, recent events, and situational requirements in an abstracted form. However, the sophistication of these environmental models varies dramatically across evolutionary history.

In evolutionarily older, non-cortical systems, these models are constrained by their architecture. Subcortical circuits model specific aspects of the environment—threat, hunger, arousal, reward—through pathways that, while capable of learning within their domain, are shaped primarily by evolutionary pressure to detect particular patterns (\cite{chin2023beyond, chalavi2018anatomy}). These systems are highly effective but domain-bound: they represent the world in specialised ways, and their scope for learning novel relational structures is limited by their fixed connectivity patterns.

Modelling the external environment alone, however, is not sufficient for consciousness. A further qualitative transition occurs when systems begin to model not only the world, but the behaviour of other internal systems.
 Instead of representing "what is out there," it begins to represent "what the system is doing." By receiving convergent input from perception, action, emotion, motivation, and reward systems, higher systems can learn the relationships between these internal processes: which perceptual states tend to trigger which actions, under which emotional or motivational contexts, and with which typical outcomes.
Importantly, these internal models emerge from the same mechanisms as all other learning—connection formation, recurrence, and reinforcement. No new cognitive faculty is required. The system simply applies its existing relational learning capabilities to a new input stream: rather than receiving photoreceptor activations or proprioceptive signals, it receives the activation patterns of other neural systems. The computational principle remains identical; only the source of input has changed. This architectural shift—from external sensory input to internal neural input—is sufficient to produce the qualitative transition from environmental modeling to self-modeling.
Through this process, causal connections that were previously only implemented implicitly become explicitly represented. The context or meta-state, that shaped a behaviour implicitly over multiple systems, now can be explicitly represented in an higher system. Higher systems do not merely react to lower systems; they learn to explain them. These explanations are then fed back through context enrichment, allowing lower systems to interpret new inputs within an abstracted internal situation model.
At this point, the meta-state no longer represents only the external environment and bodily condition. It increasingly represents the system's own internal dynamics—what it is doing and why. This transition marks the architectural basis of consciousness: the system is no longer merely responding to the world, but interpreting its own activity within it.

As additional layers accumulate, systems explain systems that explain systems, producing increasingly abstract and temporally extended self-models. Intelligence and consciousness thus emerge from the same process: the progressive expansion of relational modelling, first of the environment, and eventually of the system itself. The mammalian neocortex, as we explore next, represents the fullest realization of this architecture: a system whose uniform structure and extensive connectivity positions it to observe and model virtually all other neural systems simultaneously.

\subsection{The Role of the Neocortex: A General-Purpose Connection Maker}

The mammalian neocortex is the ultimate system explainer. Whereas older neural systems were specialized, domain-bound, and evolutionarily fixed, the neocortex emerged as a uniform, massively interconnected, and highly flexible relational system.
Its six-layered microcircuitry—remarkably consistent across all cortical regions (\cite{felleman1991distributed, markov2014anatomy})—enables one core operation everywhere: forming and refining relational connections, regardless of modality. Vision, language, motor planning, abstract reasoning—each uses the same cortical algorithm.

Importantly, the neocortex is not powerful merely because of its neuron count. Its strength lies in its connectivity. Cortical regions receive input from virtually every major system in the brain: sensory cortices, motor pathways, limbic structures, the basal ganglia, the hippocampal formation, and crucially, the hypothalamus.
It does not simply receive sensory data—it receives a complete picture of the organism’s internal landscape.
This subcortical input provides high-level cortical areas with access not only to abstract representations but also to raw affective, motivational, and homeostatic signals. As a result, cortical frameworks can integrate high-granularity sensory details with low-granularity motivational states, creating relational links that span levels of abstraction.

Because of this observational position, the neocortex learns how other systems behave. It learns that specific contextual configurations—dark forest, unfamiliar movement, elevated arousal—reliably produce specific responses in older structures: amygdala threat signals, hypothalamic stress modulation, motor preparation. Over time, the cortex internalizes these mappings. It acquires explicit causal models of the organism’s own behavioural machinery.

This learned structure enables what we might call simulation or imagination: the capacity to process situations that are not currently present. Because the neocortex has learned the causal connection between contexts and system responses, it can activate this connection without requiring the original context to be physically present. When you think about encountering a shadow in a dark forest, your neocortex activates the framework representing that context—and because it has learned what responses typically follow such contexts, it can generate or anticipate those responses as part of ongoing processing.
Crucially, we need not specify the exact mechanism by which this occurs. The neocortex might directly activate lower systems through descending connections, recreating their typical response patterns. Or it might simply represent what those systems would do, having learned their behavior well enough to predict their output without actually engaging them. What matters is that the neocortex has learned the causal mappings: context X produces response Y—and can now use these mappings internally, processing the consequences of imagined situations as if they were occurring.

This capacity emerges naturally from the systems-explaining-systems architecture. A cortical system that observes "forest-shadow context → threat-system activation → flight-preparation → relief" doesn't just learn this sequence passively. It learns the causal structure: what produces what, and why. Having learned this structure, it can activate any part of the chain and process what would follow. It can start with the context (imagining the forest), generate the expected responses (threat, flight impulse), and evaluate the outcome (safety, exhaustion)—all without external input. The system processes its own learned causal models.

Assuming the neocortex is hierarchical, with higher-level areas receiving more abstract and distilled representations, the prefrontal cortex (PFC) sits at the top of this hierarchy. Its unique connectivity places it in position to integrate the widest variety of signals—sensory, emotional, motivational, mnemonic, and abstract— and at the same time allows to influence a wide range of other systems processing. In our framework, this makes the PFC a primary highest-order system explainer: it learns the relational structure of virtually every system beneath it. This role aligns naturally with existing theories, who describe the PFC as the main faculty for intelligence (\cite{rosenthal2009higher, baars2005global, mashour2020conscious}).
In Higher-Order Thought models, the PFC carries representations about other representations.
In Global Workspace Theory, PFC-parietal networks broadcast globally influential information.
In our view, both functions emerge from the same principle: the PFC holds the most abstract, stable frameworks and, through its descending and widespread lateral connections, enriches processing throughout the brain. It is the cortical region whose frameworks most strongly influence the meta-state.

Until now, we have described these processes using the language of hierarchical organisation — higher and lower systems, ascending abstraction and descending contextual influence. It is important to emphasise that this hierarchy is a conceptual simplification rather than a literal architectural claim. Evidence increasingly suggests that cortical organisation is better described as a heterarchy, in which information flows not only upwards and downwards, but also laterally between systems of different granularity and function (\cite{hawkins2025hierarchyheterarchytheorylongrange}).
In such a heterarchical architecture, no system passively 'waits' for another to finish processing. Instead, systems continuously influence one another by exchanging context, constraints, and partial interpretations. A system explains its activity to the systems to which it is connected, and depending on the current meta-state, multiple systems can simultaneously model, modulate, or constrain one another. The appearance of hierarchical organisation emerges dynamically from patterns of interaction: at any given moment, some systems act as interpreters while others are interpreted; however, these roles are fluid rather than fixed.
This organisation enables the cortex to integrate basic sensory cues with abstract concepts, bodily needs with long-term plans, emotional states with social meaning and concrete actions with high-level goals. In this view, hierarchy is not a rigid structure, but rather a transient functional pattern that arises within a fundamentally heterarchical network of mutually explanatory systems.

Because the neocortex can integrate signals across all levels—sensory, motor, emotional, motivational, and abstract—it develops a unified internal model that explains bodily needs, perceptual patterns, behavioural tendencies, social interactions, and its own reasoning.

\subsection{Subconsciousness as Distributed Influence Within the Meta-State}
Not all active relational structures within the meta-state are interpreted by the highest-level explanatory systems. Many frameworks remain active outside reflective focus, yet they still shape perception, evaluation, and behaviour. Because every active framework contributes contextual influence through the same recursive processing loops, the system’s behaviour reflects the integrated pressure of all active structures, not only those that reach conscious interpretation (\cite{Hawkins2017sensorymotor, hawkins2025hierarchyheterarchytheorylongrange}).

Subconsciousness, in this framework, is not a separate process or a distinct cognitive domain. It is simply the set of relational structures that remain active within the meta-state but are not currently interpreted by higher-order systems. These structures provide background biases: they guide attention, shape emotional tone, activate habits, and modulate expectations. This view aligns with empirical observations that much of human behaviour is influenced by neural activity outside conscious focus (\cite{murphy2025power}).

Our account generalizes the idea proposed by Hawkins that every cortical column contributes a “vote” toward the interpretation of sensory input (\cite{hawkins2017theory}). We retain the underlying insight—that many distributed structures jointly shape processing—but broaden it beyond the neocortex. In our framework, any active subsystem—cortical, subcortical, motivational—exerts contextual influence through the recursive architecture. We avoid the term vote, which suggests a discrete combinatorial process rather than a continuous integration of relational context.

Subconsciousness thus emerges as a natural side-effect of hierarchical and heterarchical architecture. Interpretation requires a higher-order system to construct an explicit relational explanation of the pattern generated by lower systems. When such interpretation is absent, the underlying frameworks continue to influence processing implicitly. Conscious content corresponds to relational patterns that have been explicitly interpreted; subconscious content corresponds to patterns that shape behaviour without being interpreted by the highest-level systems.

Taken together, these ideas position consciousness not as a mysterious faculty, but as the natural culmination of recursively layered relational architecture. As intelligence increases, systems acquire the ability to discover increasingly complex causal relations—not only between external events, but between internal states, actions, and long-term outcomes. Recurrence allows the system to integrate relations across time; hierarchy enables higher systems to explain the activity of lower ones; heterarchy allows these explanations to shape processing across the entire brain. The neocortex, with its uniform circuitry and extensive connectivity to both cortical and subcortical systems, binds these relational structures into coherent frameworks that span multiple levels of abstraction. Consciousness arises when these explanatory frameworks become capable of interpreting their own operations and integrating this interpretation into ongoing processing. In this view, consciousness is not an addition to intelligence but the natural continuation of the same connection-forming mechanism: once a system becomes intelligent enough to recursively interpret its own relational patterns, conscious experience emerges inevitably.

\subsection{Relationship to Other Consciousness Theories}
Our framework builds on insights from several influential theories while extending them through a unified architectural account. We want to give a brief overview how our theory relates to common consciousness theories.

\textbf{Integrated Information Theory (IIT).}
IIT (\cite{tononi2016integrated})proposes that consciousness corresponds to the quantity of irreducible integrated information ($\Phi$). We share the view that consciousness requires integration across specialized subsystems. However, integration alone is insufficient. What matters functionally is how information is integrated: whether the architecture enables higher systems to learn, represent, and interpret the causal structure of lower systems. IIT characterizes the structural signature of consciousness but does not address its computational role or evolutionary origin. In our framework, integration serves a specific purpose: enabling systems to model their own operations and flexibly regulate behaviour through explicit relational structure.

\textbf{Attention Schema Theory (AST).}
AST (\cite{graziano2015attention, graziano2013consciousness, graziano2011human})argues that consciousness arises from an internal model of attention, analogous to the body schema used for motor control. This aligns with our systems-explaining-systems principle: higher systems construct models of lower subsystems to guide behaviour. However, we generalize beyond attention. Rather than modelling a single process, higher systems in our framework learn the relational patterns linking perception, motor preparation, emotional evaluation, internal drives, and reward. The attention schema becomes one instance of a broader class of explanatory schemas spanning multiple functional domains.

\textbf{Higher-Order Thought (HOT), Global Workspace Theory (GWT) and Global Neuronal Workspace (GNW).}
As we discussed, our framework is closely related to these theories. HOT (\cite{rosenthal2009higher}) emphasizes that a mental state becomes conscious when represented by a higher-order system, while GWT and GNW (\cite{dehaene2011consciousness, baars1997theatre, mashour2020conscious}) holds that conscious content is globally available for influencing diverse processes. We provide a mechanistic grounding for both insights. Higher systems receive relational patterns from lower systems and learn their structure (HOT), and these explanatory frameworks are fed back into the hierarchy via context enrichment, influencing perception, emotion, and action selection (GWT/GNW). No dedicated “workspace” or “higher-order module” is required; conscious access emerges naturally from recursive hierarchical organization that integrates structure across the entire meta-state.

Taken together, these relationships highlight that our framework does not compete with existing theories but situates their core insights within a broader architectural picture. Consciousness emerges not from a single mechanism—attention modeling, information integration, or global broadcasting—but from the recursive interaction of systems that learn, represent, and contextualize one another’s activity.

\section{Limitations}
The framework proposed in this work is intentionally conceptual. Its aim is to articulate the architectural principles that underlie intelligence and consciousness, rather than to offer a complete biophysical model of the brain. As such, several limitations must be acknowledged.\\

First, the theory does not attempt to describe the detailed neuronal and molecular mechanisms underlying connection formation, synaptic plasticity, or local circuit computation. While we refer to general architectural features—recurrent activation, hierarchical layering, heterarchical interactions between systems—these descriptions draw on established neuroscientific literature. We interpret these known mechanisms through the lens of our framework but do not claim to have tested this interpretation experimentally. This reflects both the scope of the work and the current empirical uncertainty surrounding many low-level mechanisms.

Second, many components of our framework—such as circular activations stabilizing meta-states, or higher systems interpreting relational patterns of lower ones—are consistent with existing empirical observations about cortical dynamics (e.g. \cite{salners2023recurrent, rajan2016recurrent, sherman2016thalamus}). However, we have not conducted experiments specifically designed to test whether these mechanisms function in the manner our framework proposes. Our contribution lies in providing a novel organizational lens through which existing findings can be understood, not in generating new empirical data. Future work validating or refuting this interpretive framework will require targeted experiments designed to distinguish our account from alternative explanations.

Third, the perspective offered here is necessarily incomplete. It does not cover the full diversity of neural systems, nor does it address all aspects of emotion, memory consolidation, social cognition, or language. The goal is to outline the general architecture of intelligent systems, not to provide an exhaustive account of the brain functions.

Finally, because this is a conceptual integration rather than an empirical program, we do not offer novel testable predictions that would distinguish our framework from related theories such as predictive processing, global workspace theory, or hierarchical Bayesian models. Many predictions from those frameworks remain applicable here, as we share architectural commitments. Our contribution is primarily organizational: providing a unified conceptual vocabulary (connection formation, context enrichment, systems explaining systems) that integrates insights across these traditions while offering a different functional emphasis—relational interpretation rather than prediction minimization.

These limitations reflect the status of the Stable State Framework as a high-level conceptual model. It is intended as a tool for organizing and interpreting findings across disciplines.

\section{Significance and Further Directions}

\subsection{Significance}
We believe the framework presented in this paper offers a unifying perspective on intelligence and consciousness that advances both theoretical neuroscience and the conceptual foundations of artificial intelligence.

First, the model reframes intelligence as the capacity to discover, stabilize, and use relational connections across increasingly abstract domains. By grounding cognition in connection formation and context enrichment, we provide a principled explanation for how finite biological systems achieve rich and flexible behaviour without relying on exhaustive prediction or massive sensory capacity.

Second, the framework introduces a novel account of consciousness as the natural continuation of intelligence once recursive, hierarchical, and heterarchical systems become capable of explaining the operations of other systems. Rather than treating consciousness as a separate faculty or mysterious emergent property, this perspective shows how it arises inevitably from the architecture of layered, interacting frameworks that interpret both the environment and the organism’s own activity.

Third, our emphasis on recursion and heterarchical organization highlights structural features of the brain that traditional feedforward models overlook. The ability of higher cortical regions to integrate inputs not only from subordinate cortical areas but also directly from subcortical systems (such as the hypothalamus) suggests a mechanism by which high-level abstractions can be grounded in raw sensory and motivational signals. This offers a coherent explanation for how meaning, intention, and emotion become integrated within a single meta-state.

Fourth, the framework presents a biologically inspired alternative to current AI architectures. Modern systems are powerful pattern recognizers but remain fundamentally single-system learners. Our analysis suggests that robust abstraction, compositionality, behavioural coherence, and consciousness-like properties may require multi-system relational architectures—in which separate modules learn their own patterns and higher modules learn to interpret relations among them. This offers a concrete direction for future AI research.

Finally, the relational perspective sheds new light on meaning formation and epistemic coherence. If cognitive systems seek efficient relational structure rather than explicit probabilistic accuracy, then individual and collective meaning systems depend critically on the stability and overlap of foundational frameworks. This has profound implications for understanding learning, communication, and fragmentation in modern information environments.

Taken together, these contributions form a coherent theory that links the mechanisms of learning, the structure of the meta-state, the emergence of consciousness, and the architecture of the neocortex under a single principle:
intelligence is the discovery of increasingly complex relational structure, and consciousness is what intelligence becomes when it can explain itself.

\subsection{Implications for Artificial Intelligence}
Historically, progress in artificial neural networks has followed two trajectories:
(1) increasing the number of units (\cite{rosenblatt1958perceptron}), and
(2) increasing the depth of feedforward hierarchies (\cite{lecun2015deep}).
Both dramatically expanded pattern-recognition and learning capacity.

However, modern architectures remain fundamentally feedforward single-system learners. They learn relations within one large parameter space, but lack the opportunity to understand why and how there activations influence each other.
From the perspective of the Stable State Framework, this is a central limitation.
Current architectures excel at recognizing patterns, but they cannot interpret their own activations, nor can they relate actions, rewards, and internal states across different timescales or levels of abstraction.
In essence, they lack recursion in the biological sense: the capacity to keep previous internal activations active, compare them with new ones, and discover causal connections that unfold across time.
\\
Biological intelligence relies heavily on such recursive processing, not merely on deep feedforward transformations. Recursive systems can integrate events separated by delays, observe their own behaviour, and build frameworks that relate high-level concepts to low-level sensory or motivational signals. 
 This multi-granularity relational structure is essential for abstraction, planning, causal reasoning, and the emergence of meaning.
 In artificial systems, this suggests that explicit internal reward representations may be necessary: not just scalar reinforcement signals, but functional internal states that the system can learn to treat as part of its own causal chain. Without these internal anchors, no system can truly learn the meaning of its actions.
A promising avenue for artificial systems, therefore, is not simply stacking more layers, but introducing recursive relational modules that operate on different granularities and hierarchical levels.
Such systems would more closely resemble biological intelligence and may be necessary for robust abstraction, compositionality, and long-term behavioural coherence.

This framework further suggests that general intelligence may require curriculum-like development, similar to humans, rather than training on isolated tasks. Children do not learn all concepts at once; they first acquire simple frameworks—basic shapes, movements, and social cues—supported by evolutionarily shaped circuits. These early frameworks then serve as building blocks for more complex ones, acquired through discovering new connections and compositions across contexts. This also offers an explanation for the remarkable learning speed of human cognition: we rarely construct concepts from scratch. Instead, we learn by connecting new experiences to an already rich set of pre-existing frameworks, allowing even sparse input to be interpreted efficiently. Artificial systems may therefore need to follow a comparable developmental trajectory, gradually expanding their relational world rather than attempting to learn all abstractions de novo.

A further implication concerns continual learning. Biological intelligence does not relearn the world from scratch when encountering new information; it integrates new experience into the relational structure already stabilised in the meta-state. Because the meta-state provides a persistent contextual scaffold, new learning occurs in relation to existing frameworks rather than in isolation. This prevents the kind of representational overwriting seen in many artificial systems. Instead of erasing prior knowledge, the brain adds or refines connections, guided by the higher-order frameworks that remain stable across time. In this view, continual learning is not an additional mechanism layered onto intelligence but a natural consequence of hierarchical recursion and context enrichment. Artificial systems aiming to learn continuously may therefore require analogous meta-state–like structures that preserve high-level relational context while allowing flexible adaptation at lower levels.

Finally, if recursion, relational layering, and multi-level interpretation are indeed the foundations of both intelligence and consciousness—as the systems-explaining-systems principle proposes—then building AI architectures capable of such processes may bring systems closer to consciousness-like properties. This creates both profound technical opportunities and significant ethical responsibilities. Careful consideration of these possibilities is essential before such technologies are pursued.

\subsection{Broader implications for meaning and information}

If intelligence forms meaning by searching for efficient relational structure instead of using prediction as fundamental learning mechanism, then our cognitive systems gravitate toward explanations that fit existing frameworks rather than searching for accuracy itself. Frameworks do not need to be completely accurate representations of the real world. They only need to be stable and useful enough to guide behaviour. This is generally adaptive—frameworks stabilize perception and reduce the need for constant re-evaluation—but it also implies that
incomplete information, fragmented information environments and highly selective exposure can lead to the formation of divergent frameworks across individuals or groups.
Once established, such frameworks become self-stabilizing: new information is interpreted in terms of the existing relational structure, making frameworks resistant to change.
Of course the most efficient connection between information is probably also the most accurate one, but that only holds if we receive and use all the information required, which is not always the case in nowadays information society.
While a full discussion of societal implications lies beyond the scope of this paper, the relational perspective suggests that the coherence of collective meaning systems depends critically on shared foundational connections.
A more detailed social interpretation of this idea is discussed in \cite{semmler2025efficientminds}.

\section{Conclusion}
This paper presents a conceptual framework for understanding intelligence and consciousness as emergent properties of relational structure rather than as collections of domain-specific abilities or mechanisms driven primarily by prediction. Intelligence, in this view, is the capacity of a system to form, refine, and integrate connections between signals, internal states, actions, and learned structures. Through these connections, systems construct meaning, guide behaviour, and build increasingly abstract and stable relational frameworks that support adaptive action.

Central to this framework are the concepts of the meta-state and context enrichment. The meta-state describes the system's current state as represented by its active frameworks—an internal situation model that integrates sensory input, internal state, goals, and prior knowledge. Context enrichment is the mechanism by which this internal structure shapes ongoing processing: incoming signals are interpreted through active relational frameworks rather than processed in isolation. Together, these mechanisms explain how finite biological systems achieve efficient, meaningful processing in complex environments—by reusing and integrating learned structure rather than relying on exhaustive prediction or high-throughput sensory analysis.

Building on this foundation, the paper develops the principle of systems explaining systems, grounded in recursion, as the core mechanism underlying advanced cognition and consciousness. As evolution adds layers of relational processing, higher systems learn the patterns implicitly expressed by lower systems, forming explicit models of how perception, action, emotion, motivation, and reward interact over time. Recursion allows these systems to maintain and compare internal activations across temporal scales, thus learning explicit causal connections between both external and internal events. Through context enrichment, higher-level explanations are fed back into lower systems, shaping how new inputs are processed. The meta-state reflects this unified configuration of signals and their recursively generated interpretations, enabling the system not only to model the external environment, but increasingly to model its own internal operations.

From this perspective, consciousness emerges as a natural extension of intelligence, not as an additional faculty. As systems become capable of recursively interpreting their own relational structures—understanding what they are doing, why they are doing it, and how internal processes relate to outcomes—conscious experience arises. Intelligence and consciousness are thus different expressions of the same underlying process: progressively deeper relational modelling, first of the environment and eventually of the system itself.

Although conceptual in nature, the framework has clear implications for neuroscience and artificial intelligence. It suggests that advances in general intelligence may depend less on scaling single models and more on architectures that support recursion, context enrichment, and multi-system interaction—where higher-level structures interpret and regulate lower-level processing. Such designs may enable more robust abstraction, continual learning, and coherent long-term behaviour, while also raising important questions about the emergence of consciousness-like properties in artificial systems.

By framing intelligence as connection formation and consciousness as recursive relational interpretation, this work aims to provide a clearer and more unified intuition for the architecture of cognitive systems. The framework is not proposed as a final theory, but as a foundation for further interdisciplinary exploration across neuroscience, cognitive science, and artificial intelligence—one that integrates existing insights into a coherent architectural perspective on mind and intelligence.

\section{Acknowledgments}

The author acknowledges the use of AI tools, notably Anthropic's Claude and OpenAIs ChatGPT, in the preparation of this manuscript. The AI models were utilized to assist with language refinement, formatting guidance, and organizational suggestions. The scientific ideas, theoretical development, and conclusions presented in this work are entirely the result of the author's independent research and reflection.

\bibliographystyle{unsrtnat}

\bibliography{references}  




\end{document}